\documentclass{article}
\usepackage{spconf,amsmath,graphicx,hyperref,booktabs,multirow,siunitx}

\title{On the Sensitivity of Firing Rate-Based Federated Spiking Neural Networks to Differential Privacy}

\name{Luiz Pereira, Mirko Perkusich, Dalton Valadares, Kyller Gorgônio\thanks{This study was financed in part by the Coordenação de Aperfeiçoamento de Pessoal de Nível Superior – Brasil (CAPES) – Finance Code 001.}\thanks{© 2026 IEEE. Personal use of this material is permitted. Permission from IEEE must be obtained for all other uses, in any current or future media, including reprinting/republishing this material for advertising or promotional purposes, creating new collective works, for resale or redistribution to servers or lists, or reuse of any copyrighted component of this work in other works.}}
\address{Center of Electrical Engineering and Computer, Federal University of Campina Grande, Brazil}
\begin{document}
%
\maketitle
\begin{abstract}
Federated Neuromorphic Learning (FNL) promises energy-efficient and privacy-preserving learning on devices without centralizing data. In real-world settings, deployments require additional privacy mechanisms, such as Differential Privacy (DP), whose gradient clipping and noise alter training signals. This paper analyzes how DP perturbs firing-rate statistics in Spiking Neural Networks (SNNs) and how these perturbations propagate to rate-based FNL coordination. On a speech recognition task under non-IID settings, ablations across privacy budgets and clipping bounds reveal systematic rate shifts, attenuated aggregation, and reduced client-selection stability. Moreover, we relate these shifts to sparsity and memory indicators. The results provide actionable guidance for privacy-preserving SNN-FL, highlighting the trade-offs between privacy strength and rate-dependent coordination.
\end{abstract}

\section{Introduction}
\label{sec:intro}
Federated Learning (FL) enables collaborative on-device training across resource-constrained edge clients without centralizing raw data~\cite{mcmahan2017communication,zhang2022federated}. In constrained settings, Spiking Neural Networks (SNNs) are attractive because event-driven computation can substantially reduce power, latency, and memory traffic compared to dense ANN pipelines~\cite{venkatesha2021federated,nguyen2024robustness}. Within federated SNN learning (SNN-FL), two rate-aware coordination strategies have emerged: (i) rate-weighted aggregation in asynchronous FL~\cite{wang2023efficient}, where the server mixes a client update with the current global model using a weight that also incorporates average spike rate; and (ii) active client selection~\cite{zhan2024sfedca} that prioritizes clients exhibiting large firing-rate differences to accelerate convergence under non-IID data.

In practical deployments, however, FL on user-generated data must satisfy privacy guarantees~\cite{bagdasaryan2020backdoor,collins2025federated}. The established mechanism in literature is DP-SGD: clip per-sample gradients to a bound $C$ and add Gaussian noise with multiplier $\sigma$, and privacy guarantees are tracked with accountants under subsampling~\cite{abadi2016deep,gopi2021numerical}. When training LIF-based SNNs with surrogate gradients, such perturbations do not inject noise into spikes directly but can indirectly alter the parameters controlling synaptic impulse and thresholds and, hence, the firing rates used by SNN-FL policies.~\cite{gerstner2014neuronal}.

Despite the use of firing-rate signals in SNN-FL coordination, prior studies do not quantify how DP-SGD’s clipping and noise impact: (a) the level and variance of measured rates; and (b) downstream decisions that depend on them (i.e., rate-weighted aggregation~\cite{wang2023efficient} and rate-difference client selection~\cite{zhan2024sfedca}). Our contributions are summarized as follows:

\begin{itemize}
    \item We develop a sensitivity analysis that maps DP-SGD perturbations to bias/variance in firing-rate estimators for LIF neurons and derives their propagation into (i) rate-weighted aggregation rules used in asynchronous SNN-FL and (ii) rate-difference-based client selection.
    \item Through ablations on an event-driven workload, we quantify how privacy budgets $(\varepsilon,\delta)$ and clipping $C$ reshape layer-wise firing-rate statistics and degrade (i) aggregation weights and (ii) client-ranking stability.
    We relate observed rate shifts to sparsity and memory indicators to contextualize inference-time implications.
\end{itemize}

Section~\ref{sec:background} reviews LIF/rate background, rate-aware SNN-FL formulations, and DP-SGD for FL. Section~\ref{sec:sens_analysis} presents the DP-to-rate perturbation analysis and expected consequences for rate-weighted aggregation and for rate-difference client selection. Section~\ref{sec:ablation} reports ablations, and Section~\ref{sec:conclusion} concludes the work.



\section{Background}
\label{sec:background}
\subsection{LIF Model and Firing Rate}\label{ssec:fr}

Primary computational units of Spiking Neural Networks (SNNs) are commonly modeled as a Leaky Integrate-and-Fire (LIF) neuron model~\cite{eshraghian2023training,zhang2025lif}. In continuous time, the LIF neuronal subthreshold dynamics are given by:

\begin{equation}\label{eq:lif_cont}
\tau_m \frac{dV_t}{dt} = -\big(V_t-V_{\mathrm{rest}}\big) + R\,\bar I_t,
\end{equation}
where $\tau_m$ is the membrane time constant, $V_t$ is the membrane potential, $V_{\mathrm{rest}}$ is the resting potential, and $\bar I_t := R\,I_t$ is the input current scaled by membrane resistance $R$. Let $V_{\mathrm{th}}$ denote the firing threshold and $V_r$ the reset potential. A spike $s$ is emitted whenever $V_{t^-}\ge V_{\mathrm{th}}$. Upon a spike, the state is reset $V_{t^+}\leftarrow V_r$ and held during an absolute refractory period $\tau_{\mathrm{ref}}$.

For simulation and learning, a discrete-time Euler update with step $\Delta t$ is applied to Equation~\ref{eq:lif_cont}~\cite{eshraghian2023training}. Spike generation is determined using the Heaviside step function $H(\cdot)$. Because $H(\cdot)$ is non-differentiable, gradients are computed through a surrogate derivative $\phi'(U_t-V_{\mathrm{th}})$ during backpropagation-through-time (BPTT), i.e., in the backward pass, the true derivative $H'$ is replaced by its smooth surrogate $\phi'$ (e.g., fast-sigmoid).

\textbf{Empirical Firing Rate.} Let $s_{j,t}^{(\ell)}\in\{0,1\}$ be the spike of neuron $j$ in layer $\ell$ at time $t$. For a time window of length $T$, the per-neuron firing rate is:

\begin{equation}\label{eq:per-neuron_fr}
r_{j}^{(\ell)} := \frac{1}{T}\sum_{t=1}^{T} s_{j,t}^{(\ell)} \;\in\; [0,1],
\end{equation}
measured in spikes per time. Based on Equation~\ref{eq:per-neuron_fr}, a batch-size- and neuron-weighted layer-average rate is derived to feed SNN-FL strategies~\cite{wang2023efficient, zhan2024sfedca}. For client $k$ and mini-batch index $b$ of size $B$, the layer-level rate is:

\begin{equation}\label{eq:per-layer_fr}
r_{k,b}^{(\ell)} := \frac{1}{B\,n_\ell\,T}\sum_{i=1}^{B}\sum_{j=1}^{n_\ell}\sum_{t=1}^{T} z_{k,i,j,t}^{(\ell)},
\end{equation}
where $n_\ell$ is the number of neurons in layer $\ell$. Equation~\ref{eq:per-layer_fr} definition induces a network-wide rate via neuron-weighted averages across layers, e.g., $r_k:= 1/n\sum_{\ell} \omega_\ell r_k^{(\ell)}$ with $\omega_\ell \propto n_\ell$.


\subsection{Federated SNN Learning}

Federated SNN Learning (SNN-FL) explores collaborative FL for on-device SNNs~\cite{skatchkovsky2020federated}. A standard cross-device SNN-FL setting has $K$ resource-constrained edge clients and a central server. Similar to a standard FL process, in round $r$, a subset of clients receives initial parameters, performs $E$ train local steps on private data $D_k$, and returns model updates $\boldsymbol{\Theta_{k,r}}$ to the server, which then updates the global parameters $\boldsymbol{\Theta_r}$.

In~\cite{wang2023efficient}, the server computes for each reporting client a rate-dependent coefficient—derived from the client’s average firing rate and performs an asynchronous global aggregation. In~\cite{zhan2024sfedca}, client selection is posed as a credit-assignment problem in which the credit is a function of the firing-rate difference of spiking neurons, and the $P$ top-credit clients are scheduled to participate. Both mechanisms are rate-aware yet do not account for the absence of accurate, low-variance measurements of firing rates.



\subsection{Differential Privacy for SNN-FL}\label{ssec:dp}

Using noisy gradients in first-order methods such as Stochastic Gradient Descent (SGD) has become a prominent approach for adding Differential Privacy (DP) to the training of differentiable models~\cite{collins2025federated}. At example-level DP, each edge client is noised during local training via Differentially Private Stochastic Gradient Descent(DP-SGD)~\cite{abadi2016deep}. In DP-SGD with per-sample clipping at norm $C$ and Gaussian noise multiplier $\sigma$, the noisy mini-batch gradient at local step $t$ is defined by:

\begin{equation}\label{eq:dp}
\tilde{\mathbf{g}}_t
=\frac{1}{B}\sum_{i=1}^{B}\operatorname{clip}_C\!\big(\mathbf{g}_t^{(i)}\big)
\;+\;\frac{\sigma C}{B}\,\boldsymbol{\xi}_t,\quad
\boldsymbol{\xi}_t\sim \mathcal{N}(\mathbf{0},\mathbf{I}),
\end{equation}



DP has been used in the SNN-FL setting~\cite{han2023towards,luo2025encrypted}. However, the DP mechanism in Equation~\ref{eq:dp} acts directly on surrogate gradients (see Section~\ref{ssec:fr}). Consequently, DP perturbs the learned parameters $\boldsymbol{\Theta}$ that set synaptic gains, effective thresholds, and therefore the firing-rate functionals used in SNN-FL strategies mentioned before.

\section{Sensitivity Analysis}
\label{sec:sens_analysis}

\subsection{DP Noise and Firing Rate}

Let $\boldsymbol{\Theta}$ denote all SNN trainable parameters. A parameter update can be briefly defined by $\boldsymbol{\Theta}_{t+1}=\boldsymbol{\Theta}_{t}-\eta_t\,\tilde{\mathbf{g}}_t$ in DP-SGD, t. A first-order expansion around the non-DP reference $\boldsymbol{\Theta}^\star$ yields, after $T$ private local steps,

\begin{equation}\label{eq:mean_rate}
r(\boldsymbol{\Theta}_T)
\approx
r(\boldsymbol{\Theta}^\star)
-
\Big(\sum_{t=0}^{T-1}\eta_t\Big)\nabla r(\boldsymbol{\Theta}^\star)^\top
\big(\bar{\mathbf{g}}_t^{C}-\bar{\mathbf{g}}_t\big),
\end{equation}

\begin{equation}\label{eq:var_rate}
\mathrm{Var}\big[r(\boldsymbol{\Theta}_T)\big]
\approx
\nabla r(\boldsymbol{\Theta}^\star)^\top
\Big(\sum_{t=0}^{T-1}\eta_t^2\,\frac{\sigma^2 C^2}{B^2}\,\mathbf{I}\Big)
\nabla r(\boldsymbol{\Theta}^\star),
\end{equation}
where $\bar{\mathbf{g}}_t^{\,C}=\tfrac1B\sum_i \operatorname{clip}_C(\mathbf{g}_t^{(i)})$ and $\bar{\mathbf{g}}_t=\tfrac1B\sum_i \mathbf{g}_t^{(i)}$.

Equation~\ref{eq:mean_rate} expresses a clipping bias $\boldsymbol{\delta}_{\mathrm{clip}}$ (shrinkage of large-norm directions), while Equation~\ref{eq:var_rate} quantifies the inflated dispersion of rate estimates due to injected Gaussian noise. Both effects scale into the firing-rate space through the sensitivity vector $\nabla r(\boldsymbol{\Theta}^\star)$. These approximations extend directly when the DP noise covariance deviates from $\mathbf{I}$ (e.g., per-layer clipping), by replacing $\mathbf{I}$ with the corresponding covariance.



Under noisy drive, LIF neurons exhibit rate statistics that depend smoothly on the effective mean $\mu_{\text{eff}}$ and variance $\sigma_\text{eff}$ of their input current. Random fluctuations (e.g., noisy drive) in input current $\bar I_t$ can push $V_t$ across $V_{\mathrm{th}}$ even when the noiseless trajectory would not. This “noise-assisted” spiking is a classic result for integrate-and-fire models~\cite{gerstner2014neuronal}. DP setting in this study does not inject noise into $\bar I_t$. Instead, it perturbs $\boldsymbol{\Theta}$ during training, which in turn shifts the post-training operating. For example, noise in the gradient can lead to less calibrated synapses, resulting in neurons firing less frequently. A local linearization gives a firing rate perturbation as:

\begin{equation}
\Delta r \approx \frac{\partial r}{\partial \mu_{\text{eff}}}\Delta \mu_{\text{eff}}
+
\frac{\partial r}{\partial \sigma^2_{\text{eff}}}\Delta \sigma^2_{\text{eff}}
+
\frac{\partial r}{\partial V_{th}}\Delta V_{th}
+\cdots,
\end{equation}
with $\Delta(\cdot)$ induced by the DP-SGD noise and clipping through $\Delta\boldsymbol{\Theta}$. Thus, even modest perturbations of parameters that set synaptic drive or thresholds can translate into measurable shifts in firing rates. In the next section, we analyze how this perturbation propagates to rate-weighted global aggregation and rate-difference–based client selection in federated neuromorphic learning.

\subsection{Client Selection and Global Aggregation under DP}\label{ssec:dp_cs}

In round $r$, the server samples a candidate set $N_r$ and selects the top-$P$ clients by the squared class-wise rate change~\cite{zhan2024sfedca}:

$$
\Delta R_k^r=\sum_{c=1}^{C}\!\Big(R_{k,c}(\boldsymbol{\Theta}_{k,r+1})-R_{k,c}(\boldsymbol{\Theta}_{r})\Big)^2.
$$

In~\cite{wang2023efficient}, the server assigns each reporting client a spike-rate weight

\begin{equation}
\zeta_k^r=\frac{1}{\sqrt{2\pi}\,\sigma_r}\exp\!\Big(-\frac{(r_{k,r}-\mu_r)^2}{2\sigma_r^2}\Big),
\end{equation}
and forms the asynchronous update with

\begin{equation}
\lambda_r=\kappa\,\beta_{k,r}\,\psi_k\,\zeta_k^r,\qquad 
\boldsymbol{\Theta}_r=(1-\lambda_r)\boldsymbol{\Theta}_{r-1}+\lambda_r\boldsymbol{\Theta}_{k},
\end{equation}
where $\psi_k$ accounts for sample size and $\beta_{k,r}$ for information age, $\mu_r,\sigma_r$ are round-wise statistics of clients’ spike rates computed by the server.

However, example-level DP affects the proxy and the aggregation in two coupled ways. Related to the dispersion effect, if DP increases inter-client variability of spike rates, $\sigma_r\uparrow$. Because $\zeta_k^r$ contains both the amplitude factor $1/\sigma_r$ and the discrimination term $\exp(-(\cdot)/2\sigma_r^{2})$, larger $\sigma_r$ simultaneously reduces all $\zeta_k^r$ (smaller step $\lambda_r$) and flattens relative weighting (less discrimination across clients). Conversely, a very small $\sigma_r$ makes $\zeta_k^r$ sharply peaked, concentrating the update on near-median clients and suppressing outliers. For client selection, DP-SGD can inflate and randomize the selection signal $\Delta R$, increasing the probability of ranking inversions. On the other hand, related to the center-shift effect, if DP perturbs clients toward systematically hypo/hyper-active regimes, the round center $\mu_t$ drifts. Clients whose $sr_{i,k}$ move away from $\mu_t$ are exponentially down-weighted even when the deviation is caused by DP noise rather than true data-distribution mismatch, inducing aggregation bias toward less perturbed clients and potentially harming fairness and generalization under non-IID data.

\section{Ablation Study}
\label{sec:ablation}
\subsection{Experimental Setup}

\textbf{Federated Event-based Task.} We study keyword spotting on Google Speech Commands (GSC) as an edge-relevant, event-driven task. Audio waveforms are converted to spike trains using the Speech2Spikes (S2S) algorithm~\cite{stewart2023speech2spikes}. S2S emits a fixed-length sequence of spikes per sample. Here, we use $T=200$. We load GSC via the NeuroBench framework~\cite{yik2025neurobench} with the standard train/validation/test splits. Training is federated across $K{=}10$ clients, each using its private dataset and sending model updates to a central server. All clients participate in every federated round (full participation) and contribute their locally trained models to the server.

\begin{table*}[htbp]
    \caption{Ablation under DP (means $\pm$95\% CI).}
    \label{tab:ablation}
        \centering
        \scriptsize
        \setlength{\tabcolsep}{4pt}
        \begin{tabular}{lcccccccccccccc}
        \toprule
        \multirow{2}{*}{ID} &
        \multicolumn{2}{c}{\textbf{DP config}} &
        \multicolumn{3}{c}{\textbf{Protocols}} &
        \multicolumn{6}{c}{\textbf{Metrics (mean $\pm$ CI)}} \\
        \cmidrule(lr){2-3}\cmidrule(lr){4-6}\cmidrule(lr){7-12}
        & $\varepsilon$ & $C$
        & Agg & Sel & $N/P$
        & $RMSE_{r_k}$\,($\downarrow$) & $RMSE_{r^{(\ell)}}$\,($\downarrow$) & $RMSE_{AS}$\,($\downarrow$) & $RMSE_{FP}$\,($\downarrow$) & $|\Delta\lambda|$\,($\downarrow$) & Kendall-$\tau$\,($\uparrow$) \\
        \midrule
        A0 & $\infty$ & --
           & FedAvg & All & $10/10$
           & -- & -- & -- & -- & -- & -- \\
        A1 & 8.0 & 0.5
           & FedAvg & All & $10/10$
           & \num{0.015} $\pm$ \num{0.002} & $0.054$ $\pm$ \num{0.004} & $0.015$ & $57288$ & -- & \num{0.078} $\pm$ \num{0.020} \\
        A2 & 4.0 & 0.5
           & FedAvg & All & $10/10$
           & \num{0.015} $\pm$ \num{0.002} & $0.053$ $\pm$ \num{0.004} & $0.015$ & $63467$ & -- & \num{0.059} $\pm$ \num{0.018} \\
        A3 & 1.0 & 0.5
           & FedAvg & All & $10/10$
           & \num{0.015} $\pm$ \num{0.002} & $0.052$ $\pm$ \num{0.004} & $0.015$ & $57365$ & -- & \num{0.256} $\pm$ \num{0.030} \\
        A4 & 8.0 & 0.5
           & RateW & $\Delta R$ & $10/5$
           & \num{0.015} $\pm$ \num{0.002} & $0.053$ $\pm$ \num{0.004} & $0.015$ & $63444$ & $80.780$ & $-0.268$ $\pm$ \num{0.030} \\
        A5 & 1.0 & 1
           & RateW & $\Delta R$ & $10/5$
           & \num{0.016} $\pm$ \num{0.002} & $0.052$ $\pm$ \num{0.004} & $0.016$ & $68295$ & $34.981$ & \num{0.033} $\pm$ \num{0.018} \\
        A6 & 1.0 & 2
           & RateW & $\Delta R$ & $10/5$
           & \num{0.018} $\pm$ \num{0.003} & $0.052$ $\pm$ \num{0.004} & $0.017$ & $69004$ & $19.072$ & \num{0.042} $\pm$ \num{0.019} \\
        \bottomrule
        \end{tabular}
        \vspace{2pt}
    \begin{minipage}{0.96\textwidth}\scriptsize
    \textbf{Notes.}
    $\delta=1/N$. $\alpha=1$, $K=10$, $B=64$, $E=1$, $\texttt{frac\_fit}=1.0$.
    Agg: FedAvg or RateW (rate-weighted async)\cite{wang2023efficient}.
    Sel: All or $\Delta R$ (rate-difference)\cite{zhan2024sfedca}.
    $N/P$: candidate-pool size / selected clients per round.
    RMSE$_{r}$: RMSE of client network-wide firing rate vs.\ non-DP reference (A0).
    RMSE$_{r^{(\ell)}}$: mean RMSE across layers.
    RMSE$_{AS}$: RMSE of client activation sparsity vs.\ non-DP reference (A0).
    RMSE$_{FP}$: RMSE of client footprint vs.\ non-DP reference (A0).
    $|\Delta\lambda|$: mean absolute deviation of aggregation weights from reference (A0).
    Kendall-$\tau$: stability of client ranking (non-DP vs.\ DP).
    \end{minipage}
\end{table*}

\begin{figure}[htb]
    \begin{minipage}[b]{1.0\linewidth}
      \centering
      \centerline{\includegraphics[width=8cm]{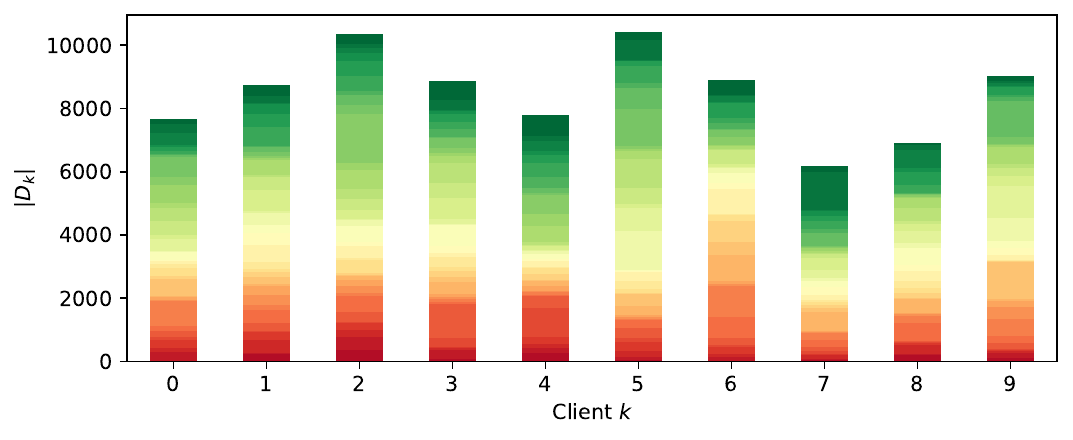}}
    \end{minipage}
    \caption{Non-IID settings on GCS dataset with $\alpha{=}1.0$. We do not describe the labels due to space constraints.}
    \label{fig:data_distribution}
\end{figure}

\textbf{Dataset under Non-IID Settings.} We adopt class-conditional Dirichlet splits with concentration parameter $\alpha{=}1.0$ to emulate realistic heterogeneity at the edge~\cite{yurochkin2019bayesian}. Client datasets are created over the GSC training set. The client data skew is reported in Figure~\ref{fig:data_distribution}. The validation dataset is partitioned similarly to the training set, and they are kept in the client for on-device metric computation. The global model is evaluated on the full test set.

\textbf{Model Architecture.} We employ a compact SNN provided by NeuroBench for the GSC task~\cite{yik2025neurobench}. We fed it with S2S spike sequences of dimensionality ($B$, $T$, 20). All LIF layers use the fast-sigmoid surrogate for gradients, threshold $V_{\mathrm{th}}=1.0$, subtractive reset, and learnable decay $\beta$ initialized at 0.9.

\textbf{Implementation Details.} We use PyTorch + snnTorch for the SNN implementation and training, Flower for FL orchestration, and NeuroBench utilities for data loading and metrics. Optimization is performed using Adam with $\eta=1e^{-3}$. Each experiment runs for 10 global rounds ($R=10$) with $B=64$ and $E=1$. After each local epoch $E$, we compute local validation metrics, including layer-wise firing rates. The server evaluates the global model at the end of each round $R$. When differential privacy is enabled, we apply example-level DP-SGD (see Section~\ref{ssec:dp}) with Gaussian noise $\sigma$ calibrated by the Privacy loss Random Variables (PRV) mechanism~\cite{gopi2021numerical} (with Poisson subsampling) to target privacy budgets $\varepsilon\in\{1,2,4,8\}$ at fixed $\delta$ and per-sample clipping norm $C\in\{0.5,1,2\}$. Non-DP runs serve as references (see Table~\ref{tab:ablation}). All experiments are conducted on an NVIDIA RTX 3050 GPU. We restrict to LIF-based SNN and a single event-driven task. Broader tasks and per-layer clipping/accounting are left for future work.


\subsection{Firing Rate Sensitivity}

We summarize the effect of the privacy budget $\varepsilon$  at fixed $\delta$ and per-sample clipping norm $C$ on rate statistics and on rate-aware SNN-FL coordination in Table~\ref{tab:ablation}.

To the FedAvg/All block (A1–A3), the network-wide rate error stays around $0.015\pm0.002$, while the layer-wise mean RMSE is $0.054{\to}0.052$ as $\varepsilon$ decreases from $8$ to $1$. Client ranking agreement with the non-DP reference is weak (Kendall–$\tau$ in $0.059{\pm}0.018$ to $0.256{\pm}0.030$), indicating that privacy noise already reshapes the $\Delta R$ signal. Client footprints (57–63k bytes) and activation sparsities remain close to A0. The rate-aware setting (A4–A6) shows the mechanism effects predicted by Section~\ref{sec:ablation}. First, the rate-weighted aggregator shows a high deviation: at $\varepsilon{=}8,\,C{=}0.5$ (A4) we observe $|\Delta\lambda|{=}80.8$, with negative Kendall–$\tau$ ($-0.268{\pm}0.030$), evidencing systematic ranking inversions. Increasing the clipping bound in string privacy $\varepsilon{=}1$ reduces the bias in the weights (A5/A6: $|\Delta\lambda|{=}35.0$ and $19.1$), but at the cost of smoothly higher variance in the rate statistics ( $\mathrm{RMSE}_{r_k}{=}0.016{\to}0.018$, $\mathrm{RMSE}_{AS}{=}0.016{\to}0.017$ ) and a also smoothly larger footprint (68 to 69k). Kendall–$\tau$ improves only marginally and remains low.

\begin{figure}[htb]
    \begin{minipage}[b]{1.0\linewidth}
      \centering
      \centerline{\includegraphics[width=8.5cm]{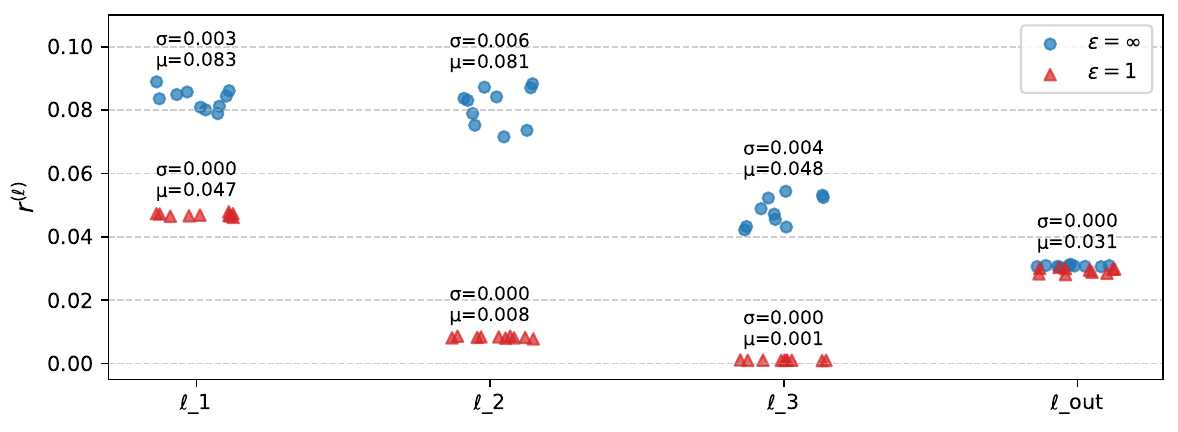}}
    \end{minipage}
    \caption{Layer-wise average firing rate $r^{(\ell)}$ as a function of the privacy budget $\varepsilon$. Markers show per-client mean across all rounds. $\varepsilon{=}\infty$ denotes the non-DP baseline. $\alpha{=}1$, $C{=}0.5$, $B{=}64$, $E{=}1$, $K{=}10$, agg=FedAvg.}
    \label{fig:layer_r}
\end{figure}

Figure~\ref{fig:layer_r} reports the layer-wise average firing rate $r^{(\ell)}$ under two privacy budgets, $\varepsilon\in\{\infty,1\}$. Hidden layers show a strong rate suppression at $\varepsilon{=}1$. Inter-client dispersion collapses with DP, while the non-DP baseline shows a small but non-negligible spread ($\sigma\in[0.003,0.006]$). The output layer is comparatively stable (non-DP $\mu\approx0.031$ vs. DP $\mu\approx 0.029{-}0.031$), indicating that DP perturbs internal activations far more aggressively than final spiking. DP-SGD’s clipping and noise shift the LIF operating point toward sparser internal activity, reducing inter-client variability.

DP-SGD’s clipping and Gaussian noise shift LIF operating points toward sparser internal activity and propagate as (i) stable, measurable drift in firing rates (A1–A3), (ii) attenuation/broadening of rate-weighted kernels that distort server mixing (large $|\Delta\lambda|$ in A4), and (iii) degraded client-ranking stability under $\Delta R$ (negative/low Kendall–$\tau$ in A4–A6).


We covers a single event-driven task and one LIF-based SNN, so DP-induced firing-rate shifts and their effects may vary. Future validation should therefore encompass multiple tasks, networks, larger client pools, and more non-IID settings. Adaptive clipping, round or client budget allocation, different accountants, and different privacy mechanisms are not explored and could change the reported bias–variance trade-offs, marking important directions for future work.

\section{Conclusion}
\label{sec:conclusion}

We studied how example-level Differential Privacy (DP) reshapes the signals that rate-aware federated coordination relies on in LIF-based Spiking Neuron Networks (SNNs). Analytically, we showed that DP-SGD’s clipping and Gaussian noise induce bias and variance in firing-rate estimates, which in turn attenuate rate-weighted aggregations and increase ranking instability for client selection based on rate differences. Ablation results in an event-driven, non-IID setting substantiate these effects and our findings suggest that, under DP, rate-dependent policies become fragile and require careful calibration if used.

\vfill\pagebreak


    %


    




\bibliographystyle{IEEEbib}
\bibliography{refs}

\end{document}